\documentclass[10pt,a4paper,conference]{IEEEtran}

\usepackage{times}
\usepackage{epsfig}

\usepackage{amsmath}
\usepackage{amssymb}
\usepackage{comment}

\usepackage{booktabs}
\usepackage{multirow}
\usepackage{arydshln}
\usepackage{graphicx, subcaption}
\usepackage{verbatim}

\begin{document}
%
\title{Let's Play Music: Audio-driven Performance Video Generation}



%
\author{\IEEEauthorblockN{Hao Zhu\IEEEauthorrefmark{1}\IEEEauthorrefmark{2},
	Yi Li\IEEEauthorrefmark{2}\IEEEauthorrefmark{3}\IEEEauthorrefmark{4},
	Feixia Zhu\IEEEauthorrefmark{1},
	Aihua Zheng\IEEEauthorrefmark{1} and
	Ran He\IEEEauthorrefmark{2}\IEEEauthorrefmark{3}\IEEEauthorrefmark{4}
	}
	
	\IEEEauthorblockA{\IEEEauthorrefmark{1}Key Laboratory of Intelligent Computing and Signal Processing of Ministry of Education, \\School of Computer Science and Technology, Anhui University, Hefei, China}
	\IEEEauthorblockA{\IEEEauthorrefmark{2}Center for Research on Intelligent Perception and Computing (CRIPAC) \\National Laboratory of Pattern Recognition (NLPR), CASIA, Beijing, China}
	\IEEEauthorblockA{\IEEEauthorrefmark{3}School of Artificial Intelligence, University of Chinese Academy of Sciences, Beijing, China}
	\IEEEauthorblockA{\IEEEauthorrefmark{4}Center for Excellence in Brain Science and Intelligence Technology, CAS, Beijing, China}

	\IEEEauthorblockA{
	Email: haozhu96@gmail.com, yi.li@cripac.ia.ac.cn, emmazfx@163.com,  ahzheng214@ahu.edu.cn, rhe@nlpr.ia.ac.cn
	}
}


\maketitle

	\begin{abstract}
	We propose a new task named Audio-driven Performance Video Generation (APVG), which aims to synthesize the video of a person playing a certain instrument guided by a given music audio clip. It is a challenging task to generate the high-dimensional temporal consistent videos from low-dimensional audio modality. In this paper, we propose a multi-staged framework to achieve this new task to generate realistic and synchronized performance video from given music. Firstly, we provide both global appearance and local spatial information by generating the coarse videos and keypoints of body and hands from a given music respectively. Then, we propose to transform the generated keypoints to heatmap via a differentiable space transformer, since the heatmap offers more spatial information but is harder to generate directly from audio. 
	\textbf{Finally, we propose a Structured Temporal UNet (STU) to extract both intra-frame structured information and inter-frame temporal consistency. They are obtained via graph-based structure module, and CNN-GRU based high-level temporal module respectively for final video generation. Comprehensive experiments validate the effectiveness of our proposed framework. }
\end{abstract}
\begin{figure*}[htb]
	\centering
	\includegraphics[width=1\linewidth]{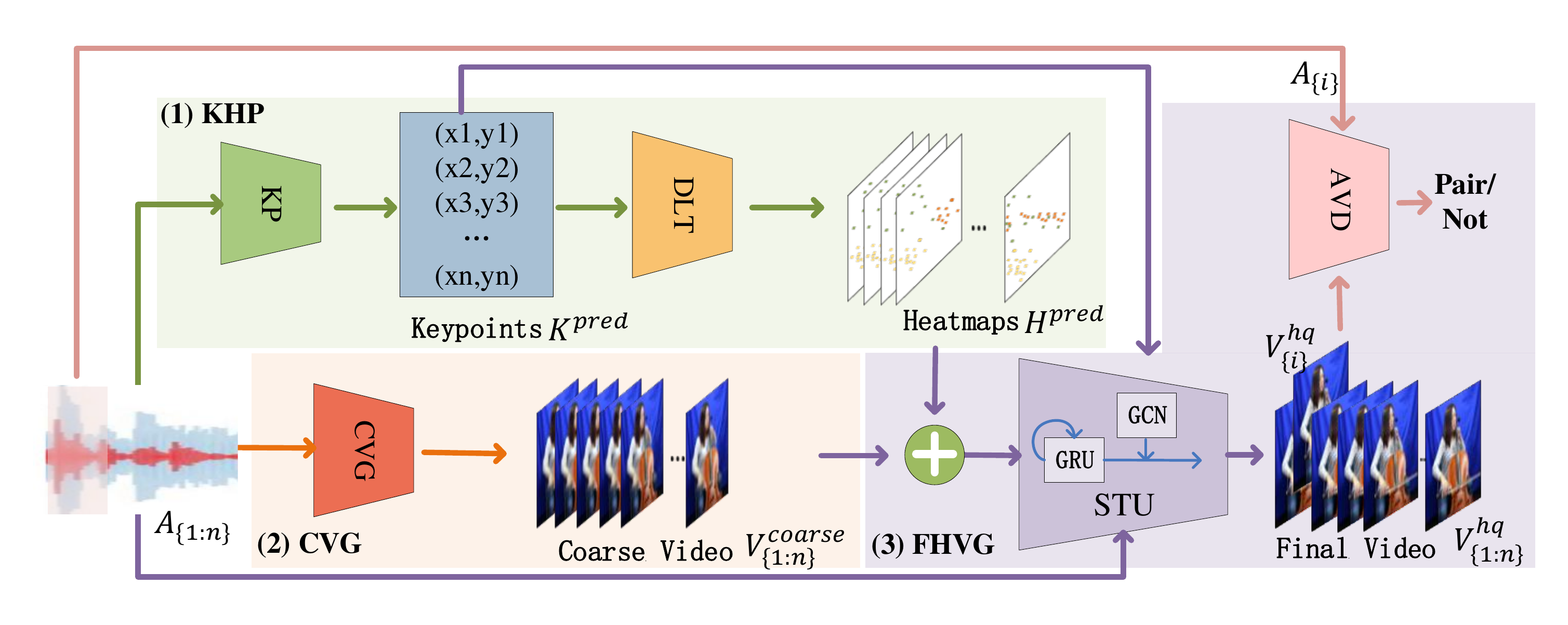}
	\caption{The pipeline of our proposed model. It contains three main steps: (1) Keypoint and Heatmap Prediction (KHP) which predicts the keypoints from the given music audio clips via Keypoints Predictor (KP), and then transforms the predicted keypoints into corresponding heatmap via Differentiable Landmark Transformer (DLT). (2) Coarse Video Generator (CVG) which generates the coarse video from given audio for further refinements. (3) Final Performance Video Generation (FPVG), which integrates the graph represented intra-frame structure information from predicted keypoints via GCN module and temporal information via CNN-GRU module.
		The representations concatenated by the generated coarse video and the predicted heatmap, the given audio via the proposed Structured Temporal UNet (STU). We finally feed the pair of each generated video frame and corresponding audio segment into the Audio Video Discriminator for judgement.}
	\label{fig:pipeline}
\end{figure*}
\section{Introduction}
\label{sec:intro}

Given a music audio of a proper instrument, professionals can distinguish which video of a certain person is playing this music, since they have the taught expert knowledge to link the relationship between the music and the corresponding performance actions. Herein, we raise a novel task in this paper: how to generate a performance video of a person playing the given arbitrary music of a specific instrument? We name this task as audio-driven performance video generation (APVG), which has widely potential applications such as concert video generation, instrumental teaching, and VR synthesis. It is a brand-new but challenging task since the extreme hardness to guide the informative motion details such as body and fingers from the heterogeneous low-dimensional audio information.

Prevalent face or body generation models employ keypoints or heatmap to guide the generation \cite{song2018geometry,songls2018geometry,chan2018everybody,ma2017pose,jalalifar2018speech,chen2019hierarchical}. Specifically, the heatmap achieves more impressive performance by offering more spatial information \cite{ma2017pose,song2018geometry,songls2018geometry,chan2018everybody}. However, keypoints are much easier to predict from audio \cite{jalalifar2018speech,chen2019hierarchical} since the heatmap is generally sparse and tends to introduce blurry and jittery generation. 
In order to utilize the advantages of both keypoints and heatmap, we propose to transform keypoints to the heatmap via a differentiable space transformer inspired by \cite{li2019layoutgan}, then use the informative heatmap to guide the body generation in the performance videos. 

Furthermore, conventional works leverage keypoints as a condition to guide the audio-driven video generation \cite{kumar2017obamanet,chen2019hierarchical}, which have ignored the rich structure information in the coordinates layout of the keypoints. In order to explore the local structure information during the audio-driven body generation, we further utilize Graph Convolutional Network (GCN) \cite{kipf2017semi}, which is one of the prevalent method to encode discrete features with intrinsic structure, to discover the intra-frame structured information from feature blocks. 

The key issue of AVPG is to generate temporally smooth performance frames.
Conventional video generation schemes either lack of temporal information \cite{chan2018everybody} or leverage computational optical flow to discover the temporal information \cite{wang2018vid2vid}. 
UNet \cite{ronneberger2015u}, which utilizes skip-connections to pass the features to the corresponded decoder layer, has been noted as a prevalent architecture with promising performance in image-to-image tasks \cite{balakrishnan2018synthesizing,isola2017image}. However, the conventional UNet cannot capture the temporal information, which is crucial in video generation.  Recently, GRU (Gated Recurrent Unit)~\cite{cho2014learning} has been drawn increasing attention in computer vision tasks due to its ability of providing long term memory of previous frames. Therefore, we propose to concatenate UNet in adjacent frames by propagating high-level feature of current frame to next frame via CNN-GRU to preserve the inter-frame temporal consistency during generation. 
Conventional GRU leverages FC layers to capture the temporal information while destroying the spatial information in original image space. CNN-GRU replaces the FC layer by Conv layer to preserve the spatial information in high-level features and achieves better performance in practice \cite{chen2019hierarchical}. 

Based on above discussion, we propose a multi-stage approach to capture both intra-frame spatial structure and inter-frame temporal consistency for audio-driven performance video generation. The overall architecture and the pipeline of our method is illustrated in Fig.~\ref{fig:pipeline}. 
To the best of our knowledge, this is the first work exploring the audio-driven performance video generation (APVG) task. The main contributions of this work can be summarized as:
\begin{itemize} 
	\item We propose an effective multi-stage adversarial generation model to achieve the APVG task, which casts a new challenging problem for audio-visual computation and provides a baseline framework for related researches and potential applications.
	\item We propose to transform the predicted keypoints to corresponding heatmap by utilizing a differentiable landmark transformer (DLT) to provide more precise local spatial information, followed by the concatenation with the coarse video generated by the given music clips, to provide global appearance information for APVG.
	\item We propose an Structured Temporal UNet (STU)  for the high-quality performance video generation in APVG, which can simultaneously capture the intra-frame structure information via graph-based representation on the predicted keypoints and inter-frame temporal consistency via CNN-GRU connected UNet.
\end{itemize}

\begin{figure*}[h]
	\centering
	\includegraphics[width=1\linewidth]{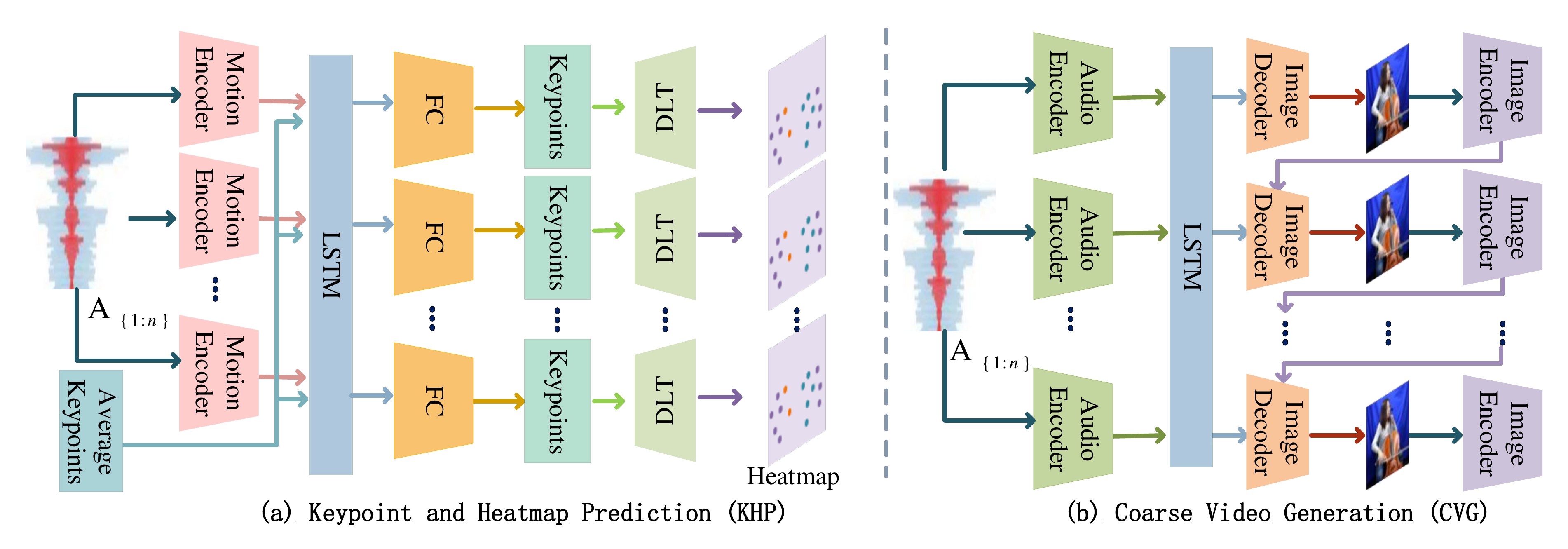}
	\caption{The pipeline of our proposed (a) Keypoint and Heatmap Prediction (KHP), and (b) Coarse Video Generation (CVG).}
	\label{fig:2architectures}
\end{figure*}

\section{Related work}

Although APVG is a new task in audio-visual generation, there are some similar tasks such as: music-driven pose motion synthesis, talking face generation and human pose transfer.

\subsection{Music-driven Pose Motion Synthesis}
Given the audio music, music-driven pose motion synthesis aims to predict a sequential structure of the body followed rendering or avatar animation to produce the final motion videos.
\cite{fan2011example} explored the relationship between the music and motion by training a music-motion matching quality rating function. 
\cite{alemi2017groovenet} proposed a real-time GrooveNet based on Conditional Restricted Boltzmann Machines (FCRBM) and Recurrent Neural Networks (RNN) to generate dance movement from music. 
\cite{lee2018listen} proposed to leverage an auto-regressive encoder-decoder network to generate choreography system from music.
\cite{shlizerman2018audio} proposed to generate keypoints of the body from audio, followed by the avatar animation.  
Recently, %
Zhuang et al. \cite{zhuang2020music2dance} leveraged global and local feature to shift the WaveNet \cite{oord2016wavenet} from speech generation to the pose motion synthesis.
Lee et al. \cite{lee2019dancing2music} decompose a dance into dance units, and proposed a network to learn how to reorganize these units via given music. 
However, these methods mainly synthesize the keypoints or skeletons to describe the body motion then generate the motion video by the renderer, while our APVG task directly generates the body motion videos from music.

\subsection{Talking Face Generation.} Given a audio clip, talking face generation aims to synthesize a realistic talking face video with lip synchronization of facial motion over the entire video speech. Earlier works synthesized talking face for a specific person \cite{kumar2017obamanet,suwajanakorn2017synthesizing}, while most recent methods focus on the synthesis for arbitrary identity \cite{Zhou2018TalkingFG,zhu2018high,chen2019hierarchical}. \cite{Zhou2018TalkingFG} proposed to disentangle the audio-visual representation into word-related and identity-related representation. 
\cite{zhu2018high} introduced mutual information approximation to capture high-level coherence between audio and visual modalities. 
\cite{chen2019hierarchical} transferred audio to facial landmarks and then generating attention and motion masks on the landmarks for final video frames. 
However, the task of synthesizing the global body motion, together local finger motion from the given audio, is more challenging and complex.

\subsection{Human Pose Transfer}
Human pose transfer aims to generate the image of a person in arbitrary poses. 
This task was first proposed by Ma et al. \cite{ma2017pose} which leveraged the coarse-to-fine scheme to synthesize the target person from the heatmap obtained from 18 keypoints. 
Balakrishnan et al. \cite{balakrishnan2018synthesizing} divided pose transfer problem into several sub-tasks and synthesized the target foreground and background separately to adapt the complex background scenes. 
UNet based architecture is a prevalent approach for pose transfer task, while hard to apply for non-aligned objects. Siarohin et al. \cite{siarohin2018deformable} introduced deformable skip connections to GAN to handle the non-aligned input and output. 
Pumarola et al. \cite{pumarola2018unsupervised} further proposed a fully unsupervised pose generation scheme by mapping the original pose image back from the generated one via a bidirectional generator. 
However, they mainly devoted to synthesize the high quality image of a person in different poses while lacked of temporal information.

\section{Approaches}

\begin{figure*}[htb]
	\centering
	\includegraphics[width=1\linewidth]{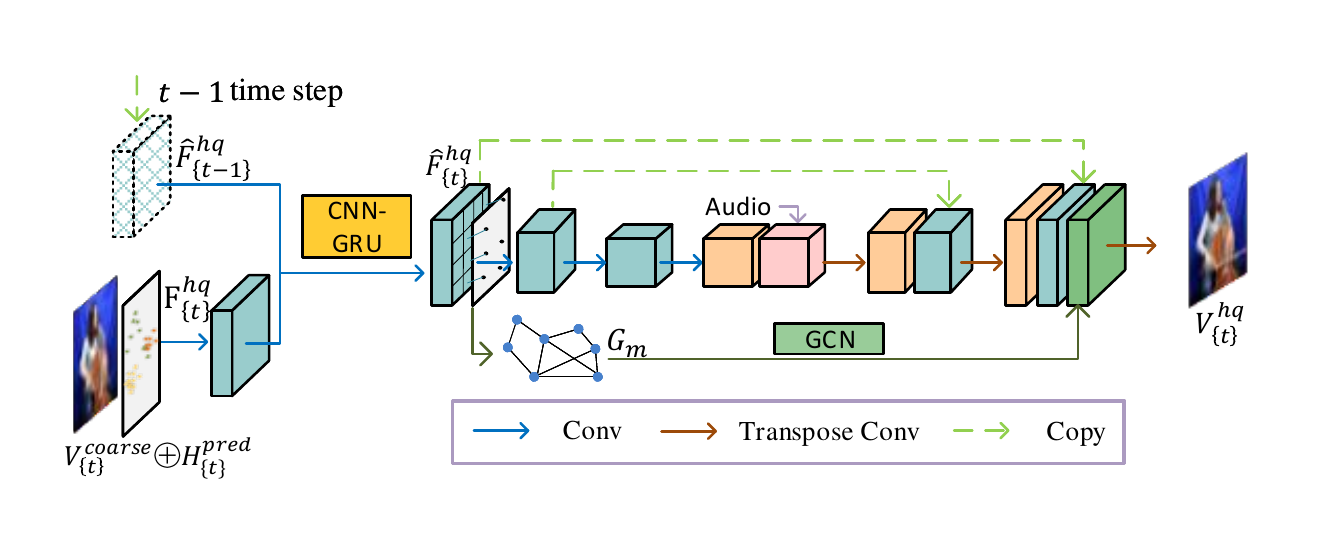}
	\caption{Illustration of our proposed STU. In  the $t$-th time step, we first extract $F^{hq}_{\{t\}}$ from $V^{coarse}$ and ${H}^{pred}$, then  fuse $F_1^t$ and $F^{hq}_{\{t-1\}}$ by CNN-GRU which produces $\hat{F}^{hq}_{\{t\}}$. Second, we use $K^{pred}$ to construct adaptive graph $G_m$ via $\hat{F}^{hq}_{\{t\}}$, then pass $G_m$ to the GCN to extract motion-related information. Finally, we extract audio feature via decoder to concatenate with the first layer of decoder, then fuse all the extracted features of same level and propagate to higher resolution layers.}
	\label{fig:gemunet}
\end{figure*}
Given an audio sequence $A_{\{1:n\}}$ ($n$ denotes the number of the audio clip) which contains the music of a proper instrument, our purpose is to synthesize a high-quality performance video $V^{hq}$. Our method consists of three parts: Keypoint and Heatmap Predictor (KHP), Coarse Video Generation (CVG), and Final High-quality Video Generation (FHVG) As shown in Fig \ref{fig:pipeline}, we shall elaborate each part in this section.

\subsection{Keypoint and Heatmap Prediction: KHP}
\label{sec:keypoint_generation}
To take both advantage of keypoints (easy to predict) and heatmap (with more spatial information) during the body motion generation, we propose to first predict the keypoints from the given audio via Keypoint Predictor (KP), and then transform the predicted keypoints into corresponding heatmap via Differentiable Landmark Transformer (DLT) for further video generation. 
Based on the prior experiments that, people act in different motion templates while playing different instruments, we consider two parts features in keypoints prediction: (1) instrument-related feature, to determine the approximate position of predicted keypoints by feeding the average keypoints from the training set into the keypoints predictor. (2) motion-related feature, to predict more precise positions extracted from Motion Encoder (1D-CNNs) on the current audio clip. We concatenate these two features and feed them to LSTM and FC layer to predict the keypoints. 

Then, we transform predicted keypoints $K^{pred}$ to corresponded heatmap $H^{pred}$ via the DLT inspired by \cite{li2019layoutgan}. $H^{pred}$ is first filled with the scalar value 0, then calculated with the following equation: 
\begin{equation}
\begin{split}
H^{pred} = \sum_{i=0}^{W}\sum_{j=0}^{H}\sum_{s=0}^{P}&\alpha*max(0,1 - |H^{pred}_i - K^{pred}_{s_x}|) 
\\& *max(0,1 - |H^{pred}_j - K^{pred}_{s_y}|), 
\label{eq:differetiable_mapping}
\end{split}
\end{equation} 
where $W$, $H$, and $P$ denote the width and height of the image and number of keypoints respectively. 
Furthermore, $s_x$ and $s_y$ denote the x-axis and y-axis coordinate of $s$-th keypoint and $\alpha = 1$ is an intensity factor. The real heatmap $H^{real}$ can be obtained in the same manner. 
The full pipeline is illustrated as Fig. \ref{fig:2architectures} (a).

During training stage, we first apply $L2$ loss between predicted keypoints $K^{pred}$ and real keypoints $K^{real}$ in Cartesian coordinate space:
\begin{equation}
\mathcal{L}^{Kpts}_{Coor} = \parallel K^{pred}_{\{1: P\}} - K^{real}_{\{1 : P\}}\parallel_2, 
\label{eq:kpts_l1}
\end{equation}

We calculate the second loss in visual space with L1 loss:
\begin{equation}
\mathcal{L}^{Kpts}_{Vis} = \parallel H^{pred} - H^{real}\parallel_1.
\label{eq:kpts_img}
\end{equation} 

By applying the two losses in (Eq. \eqref{eq:kpts_l1} and Eq. \eqref{eq:kpts_img}), we can obtain the loss of the predicted keypoints in both Cartesian coordinate space and spatial visual space which improve the keypoints prediction and facilitate the further video generation.

\subsection{Coarse Video Generation: CVG}
Despite of the local spatial information, the global appearance information, which can maintain the context of the video, is also crucial in generation. 
Therefore, we propose a coarse video generator (CVG) to simultaneously generate the general body appearance within each frame and smooth transition between adjacent frames from give music clip. 

As shown in Fig. \ref{fig:2architectures} (b), CVG consists of an AudioEncoder, an ImageEncoder, and an ImageDecoder. AudioEncoder processes audio sequence $A_{\{1:n\}}$ into audio features $F^a_{\{1:n\}}$ then feed to LSTM to obtain temporal information. ImageEncoder contains the top five layers of pretrained VGG network \cite{simonyan2014very} and two additional convolution layers. 
In order to improve the continuity in the motion, we feed previous generated frame to the ImageEncoder to extract image feature $F^v_{\{t-1\}}$.
Finally, we concatenate $F^a_{\{t\}}$ and $F^v_{\{t-1\}}$ along with a random variable $z$ to ImageDecoder to obtain the current coarse video frame $V^{coarse}_{\{t\}}$. 

Since we only expect the coarse video generation at this stage, we simply employ $L1$ loss between the real video frame $V^{real}_{\{t\}}$ and the generated coarse video frame $V^{coarse}_{\{t\}}$ for reconstruction:
\begin{equation}
\mathcal{L}^{vid}_{coarse} = \frac{1}{n}\sum_{t=1}^{n}\parallel V^{real}_{\{t\}}-V^{coarse}_{\{t\}} \parallel_1.
\label{eq:s1_l1}
\end{equation}

The output coarse videos can provide the general appearance information for the final high-quality video generation. Therefore, concatenate generated $V^{coarse}$ and ${H}^{pred}$ to feed to the next stage. 

\subsection{Final High-quality Video Generation: FHVG}
\begin{figure*}[htb]
	\centering
	\begin{subfigure}[b]{0.29\linewidth}
		\includegraphics[width=\linewidth]{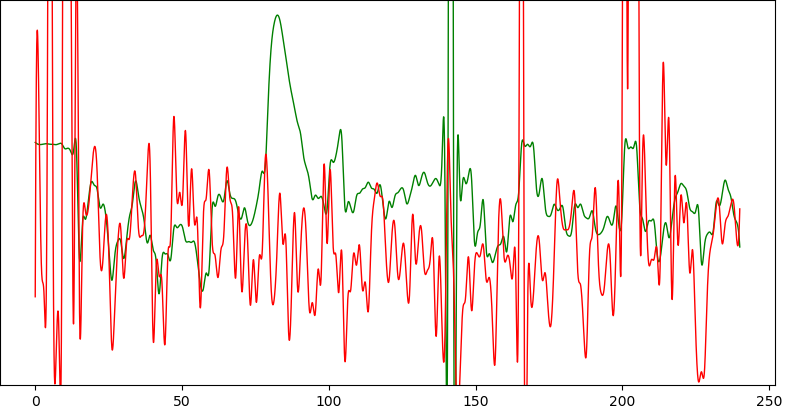}
		\caption{Baseline: left hand}
	\end{subfigure}
	\begin{subfigure}[b]{0.29\linewidth}
		\includegraphics[width=\linewidth]{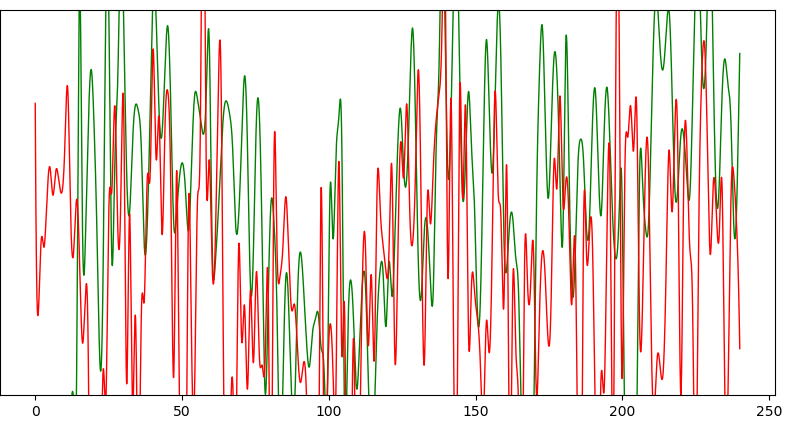}
		\caption{Baseline: right hand}
	\end{subfigure}
	\begin{subfigure}[b]{0.29\linewidth}
		\includegraphics[width=\linewidth]{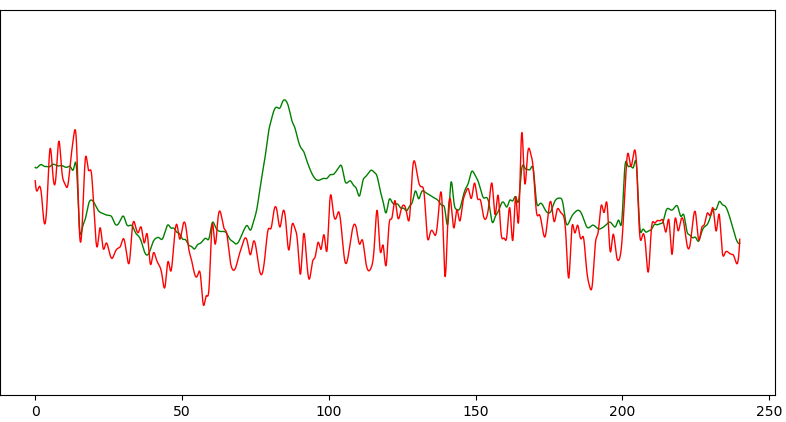}
		\caption{Baseline: body}
	\end{subfigure}
	
	\begin{subfigure}[b]{0.29\linewidth}
		\includegraphics[width=\linewidth]{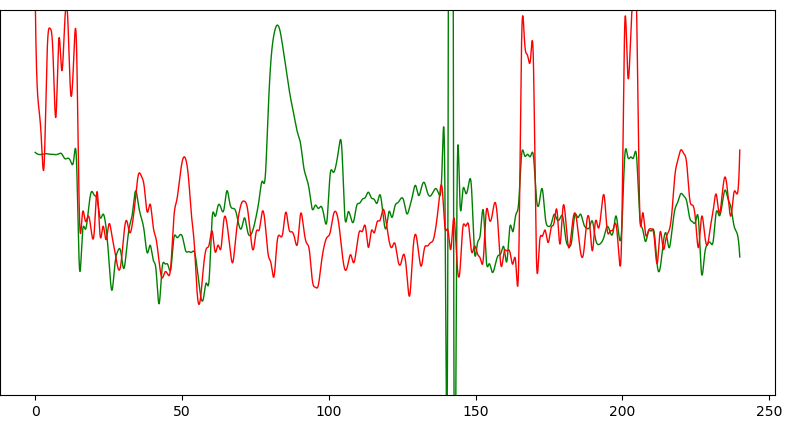}
		\caption{Ours: left hand}
	\end{subfigure}
	\begin{subfigure}[b]{0.29\linewidth}
		\includegraphics[width=\linewidth]{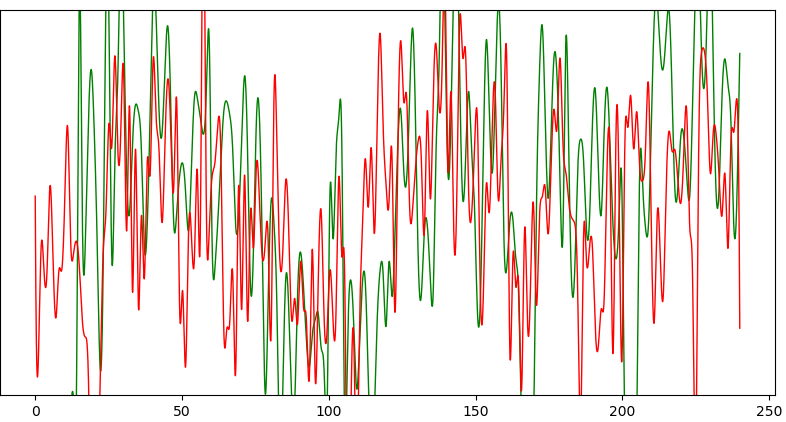}
		\caption{Ours: right hand}
	\end{subfigure}
	\begin{subfigure}[b]{0.29\linewidth}
		\includegraphics[width=\linewidth]{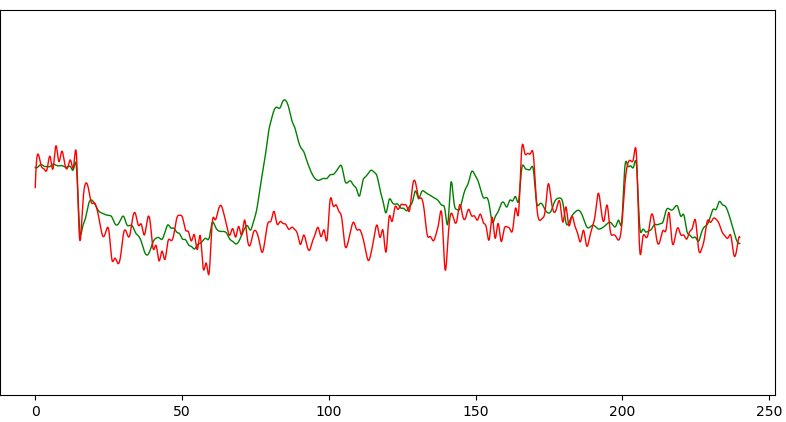}
		\caption{Ours: body}
	\end{subfigure}
	\caption{Visualization of cello keypoints, where X-axis and Y-axis denote each sample and the 1-D PCA feature respectively. The red line and green line indicate the PCA features of predicted and ground truth keypoints respectively.}
	\label{fig:keyponts_pca}
\end{figure*}

To capture both intra-frame structure information and inter-frame temporal consistency, we propose a Structured Temporal UNet (STU) by leveraging the middle level information (the predicted keypoints $K^{pred}$ and generated coarse video $V^{coarse}$) for final high-quality video generation, as shown in Fig. \ref{fig:gemunet}.

Firstly, we employ UNet \cite{ronneberger2015u} as our basic network, which is a prevalent network in image-to-image translation due to its ability of propagating context features from lower layers to higher resolution layers. However it ignores inter-frame temporal consistency, and suffers a jitter problem while synthesizing videos~\cite{wang2018vid2vid}. Herein, we propose to further temporally propagate a high-level feature between adjacent frames through the gated unit, then obtain the fused feature similar as GRU \cite{chung2014empirical}, but replacing the FC layers by CNNs to preserve spatial information. We refer to it as CNN-GRU in our paper. 

Furthermore, conventional UNet only contains CNNs to extract features in spatial-level, while neglecting the intrinsic structure information. Therefore, we propose to explore the intra-frame structured between the motion components (the feature blocks located by the predicted keypoints) via GCN due to its ability of encoding the discrete features with the intrinsic structure.

The graph of motion components can be represented as $G_m = (\mathcal{V}, \mathcal{E}, \mathcal{A})$, where $\mathcal{V}, \mathcal{E}, \mathcal{A}$ denote the nodes, edges, and adjacency matrix of the graph respectively. The nodes of the graph are the feature blocks $\hat{F}^{hq}_{\{t\}}$ located by keypoints coordinates, and the edges are connected in the same manner as performed in OpenPose \cite{cao2018openpose}. Then we feed $G_m$ into GCN to aggregate this intra-frame local features to preserve the structure relationship during final generation. 

Finally, we feed the keypoints, the heatmap concatenated coarse video, together with the given audio into the proposed STU to capture both intra-frame structure and temporal consistency for final video generation.
An additional Audio-Video Discriminator is introduced to distinguish whether the given audio and video are paired. The STU and Audio-Video Discriminator therefore formed as a GAN \cite{goodfellow2014generative}, STU tries to fool the discriminator while the discriminator attempts to find the unpaired audio and video frames. The adversarial loss is:
\begin{equation}
\mathcal{L}^{G}_{hq} = \mathbb{E}[log(D(G(A, V^{coarse}, K^{pred}),A)],
\label{eq:s2_gan}
\end{equation}
and discriminator is trained with:
\begin{equation}
\begin{split}
\mathcal{L}^{D}_{hq} = &\mathbb{E}[log(D(V^{real},A)]+\\
&\mathbb{E}[log(D(G(A, V^{coarse}, K^{pred})), A)].
\end{split}
\label{eq:s2_d}
\end{equation}

Instead of simply using L1 loss in coarse video generation, we use perceptual loss \cite{johnson2016perceptual} to capture high-level differences between the generated and real videos:
\begin{equation}
\mathcal{L}_{hq}^{perc} = \frac{1}{n}\sum_{i=1}^{n} \parallel \psi(V^{real}_{\{i\}})- \psi(V^{hq}_{\{i\}}) \parallel_1,
\label{eq:s2_perc}
\end{equation}
where $\psi$ denotes the output of different VGG-19 layers. 

\section{Experiments}
We evaluate our model on Sub-URMP \cite{chen2017deep} dataset to demonstrate the effectiveness of our proposed method for APVG task, followed by a detailed  ablation study on each component and comparing our STU against other state-of-the-art video-to-video generation models.

\subsection{Dataset and Implementation Details.}

{\bf Sub-URMP \cite{chen2017deep} } dataset consists of 13 instrument categories. Each category includes the performance videos of music clips recorded by 1 to 5 different people.
In our experiments, we choose $cello$ and $trombone$ categories which contain 8000+ frames per person in the training set. We crop each frame into a square and resize to 256*256. Audios are extracted into Constant-Q transform (CQT) features~\cite{schorkhuber2010constant} at the sampling rate of 44100Hz and hop length of 256 while each feature has a size of 84*87. 


We adopt Adam optimizer with the learning rate starting from 0.001 and then gradually decreasing to 0.000125 during training. All the parameters in networks are initialized with Kaiming initialization \cite{he2015delving}.

\begin{figure*}[htb]
	\centering
	\includegraphics[width=1\linewidth]{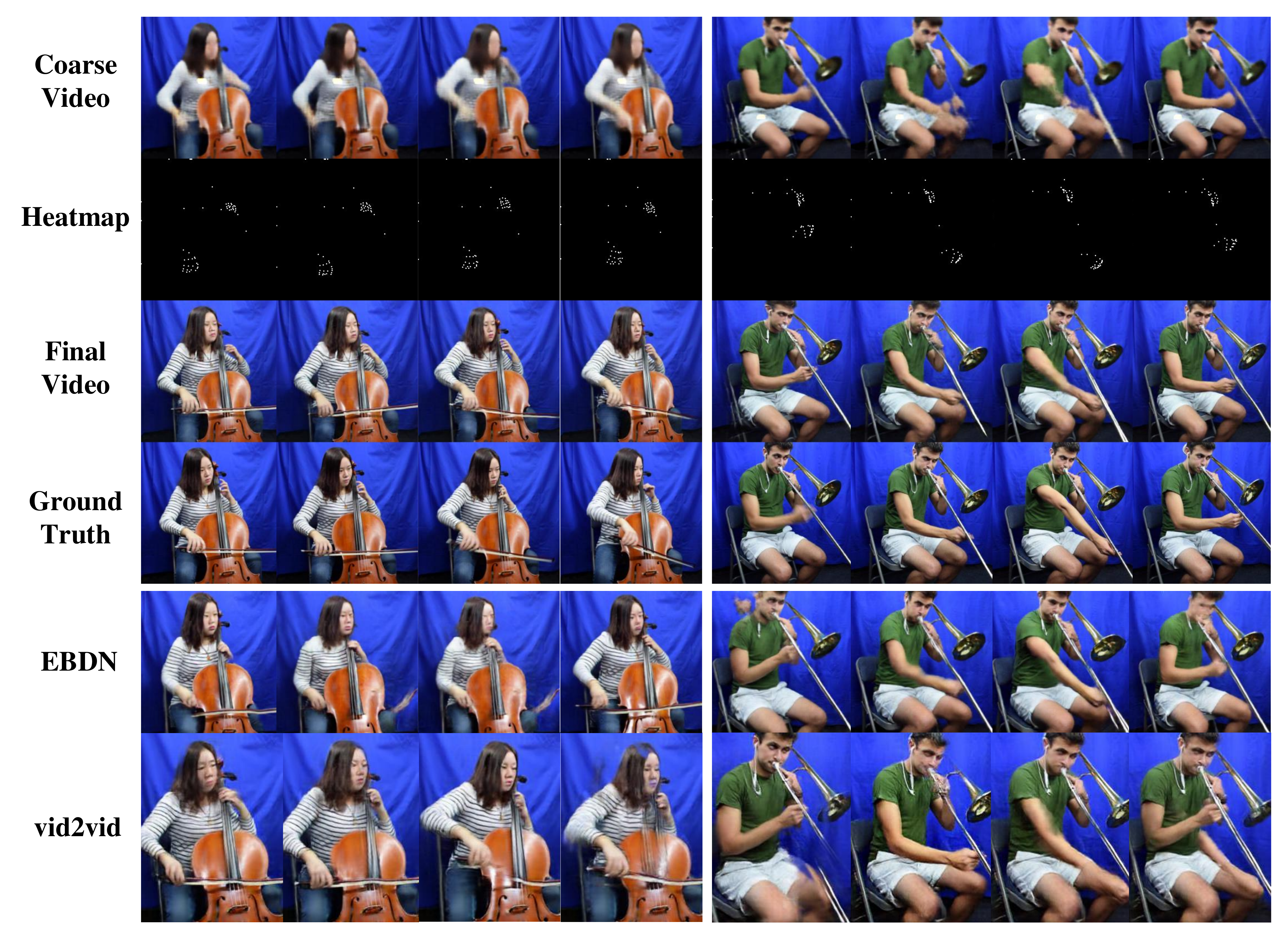}
	\caption{The generation examples of our model. Note that the results of vid2vid \cite{wang2018vid2vid} was with the size of 144*256, which was resized into 256*453 for better formatting.}
	\label{fig:quality-samples_cello}
\end{figure*}

\subsection{Ablation Study}
We first evaluate the contribution of each component in our method. We evaluate the qualitative result by the prevalent metrics: Peak Signal to Noise Ratio (PSNR), Structure Similarity Index Measure (SSIM) \cite{wang2004image}. Table \ref{tbl:video-ablationstudy} reports the ablation study result, from which we can see, (1) All the three components, KHP, CVG and STU play important roles in our method. (2) By removing the GCN model (d) or CNN-GRU module (e) from STU, both PSNR and SSIM increase, which indicates their contributions.

\begin{table}[tb]
	\centering
	\caption{Quantitative evaluation of the proposed performance video generation. (Ours = Baseline  + KHP+ CVG + STU, STU = GCN + CNN-GRU. )}
	\label{tbl:video-ablationstudy}
	\begin{tabular}{lcccc} 
		\hline
		\multicolumn{1}{c}{\multirow{2}{*}{Methods}} & \multicolumn{2}{c}{Cello} & \multicolumn{2}{c}{Trombone} \\ 
		\cline{2-5}
		\multicolumn{1}{c}{} & \multicolumn{1}{l}{PSNR} & \multicolumn{1}{l}{SSIM} & \multicolumn{1}{l}{PSNR} & \multicolumn{1}{l}{SSIM} \\ 
		\hline
		(a) w/o CVG & 15.073 & 0.306 & 13.563 & 0.206 \\
		(b) w/o KHP & 15.191 & 0.465 & 14.753 & 0.305 \\
		(c) w/o STU & 15.767 & 0.536 & 14.656 & 0.362 \\
		(d) w/o GCN & 16.253  & 0.551 & 15.572 & 0.395 \\
		(e) w/o CNN-GRU & 16.437 & 0.548 & 15.519 & 0.395 \\\hline
		Ours & \textbf{17.073} & \textbf{0.563} & \textbf{15.910} & \textbf{0.397} \\ 
		
		\hline
	\end{tabular}
\end{table}

\subsection{Evaluation on Keypoints Predictor (KP)}
\label{sec:exp_keypts}

To evaluate the performance of our keypoints prediction. We calculate the L2 distance between the predicted keypoints and the real keypoints of the proposed KP together with its two variants	
\begin{table}[tbh]
	\centering
	\caption{Evaluation different components in keypoints prediction. (Ours = Baseline  + $avg.$ condition + DLT)}
	\label{tbl:keypts-ablationstudy}
	\begin{tabular}{lccc} 
		\hline
		\multicolumn{1}{c}{\multirow{2}{*}{Methods}} & \multicolumn{3}{c}{Mean Keypoints Distance} \\ 
		\cline{2-4}
		\multicolumn{1}{c}{} & Cello & Trombone & Mean \\ 
		\hline
		Baseline & 0.164 & 0.598 & 0.381 \\
		+ $avg.$ condition & 0.151 & 0.427 & 0.289 \\
		+ DLT & 0.152 & 0.490 & 0.321 \\\hline
		Ours & \textbf{0.117} & \textbf{0.392} & \textbf{0.254} \\
		\hline
	\end{tabular}
\end{table}
\begin{figure*}[htb]
	\centering
	\begin{subfigure}[b]{0.49\linewidth}
		\includegraphics[width=\linewidth]{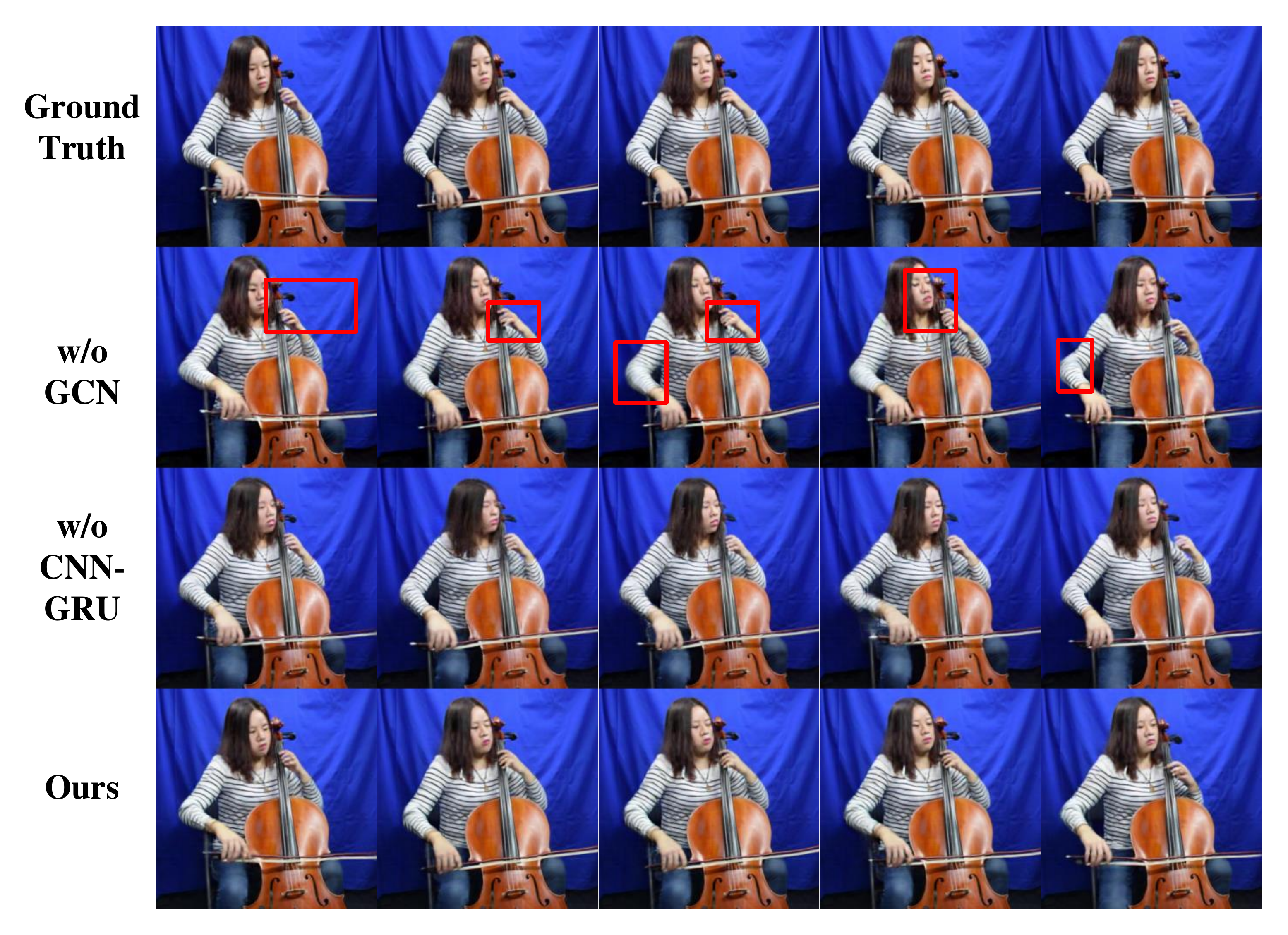}
		\caption{Comparison of cello results.}
	\end{subfigure}
	\begin{subfigure}[b]{0.49\linewidth}
		\includegraphics[width=\linewidth]{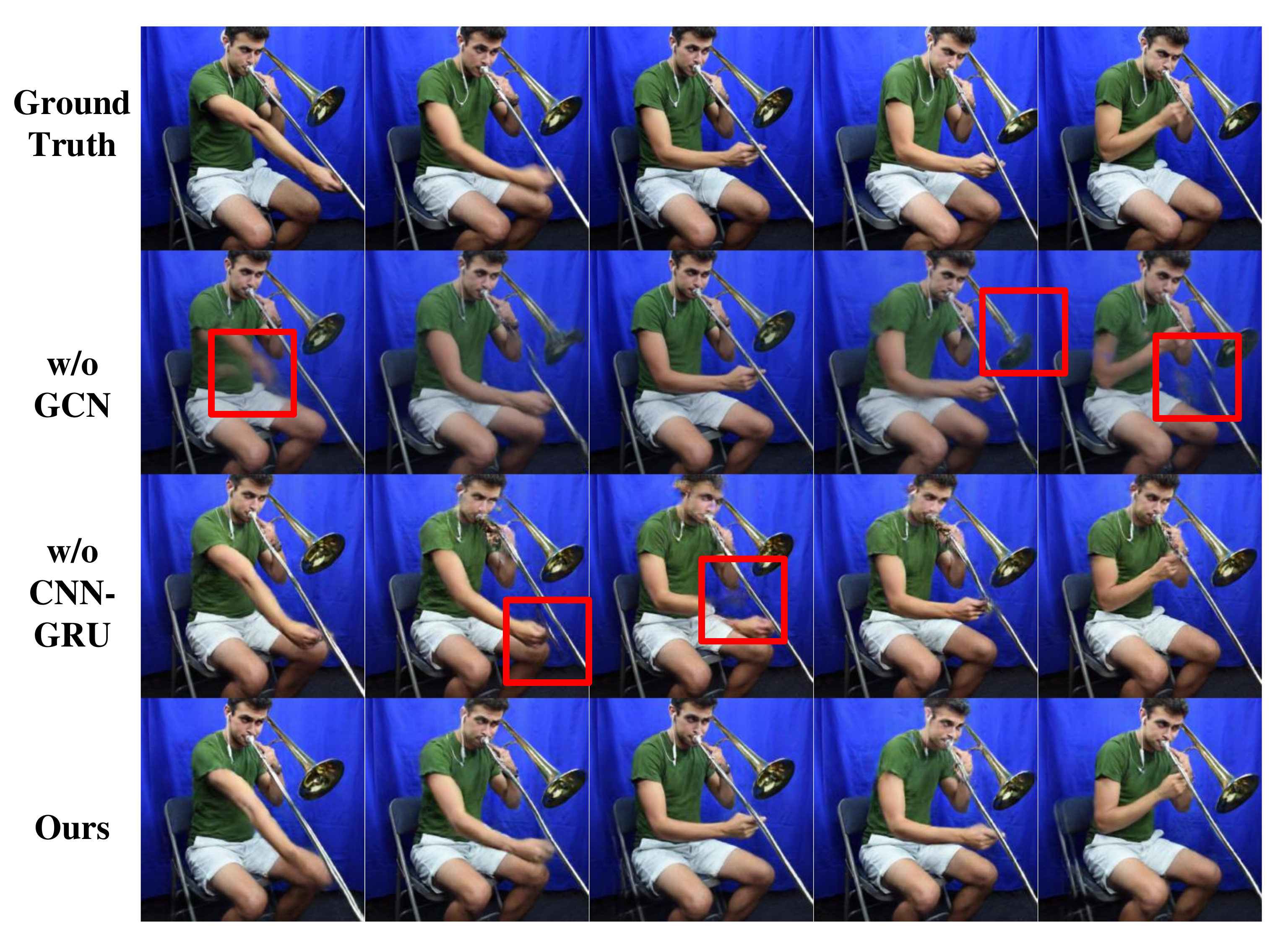}
		\caption{Comparison of trombone results.}
	\end{subfigure}

	\caption{The quality of generation with different experimental setting.}
	\label{fig:supplymentarysamplecello}
\end{figure*}
As reported in Table \ref{tbl:keypts-ablationstudy},  the distance increased after removing the average condition ($avg.$ condition) or the differentiable landmark transformation (DLT), which verifies the contribution of each component. 
We further notice that the distance of trombone videos is much larger than that of cello, the reason is that OpenPose \cite{cao2018openpose} fails more frequently to detect the ground truth keypoints on Trombone videos than on Cello ones, which affects the final video generation. 

Fig. \ref{fig:keyponts_pca} visualizes the turbulence between our predicted keypoints and the ground truth comparing to the baseline. It is clearly that our keypoint predictor can predict smoother and preciser keypoints than baseline (without average condition and the differentiable transformation). Note that the extremely large or small ground truths indicate the failure detection of OpenPose~\cite{cao2018openpose}. 			

\subsection{Evaluation on Structured Temporal UNet (STU)}

As the first task of performance video generation, we leverage the predicted heatmap (transformed from landmark) as input and compare our proposed STU (video-to-video) with other state-of-the-art video generation methods,  EveryBody Dance Now (EBDN) \cite{chan2018everybody} and vid2vid \cite{wang2018video}. As reported in Table \ref{tbl:video-comparsion}, our STU significantly beats the state-of-the-art video generation methods in all metrics. Fig. \ref{fig:quality-samples_cello} further demonstrates two comparison examples on cello and trombone categories respectively. From Fig. \ref{fig:quality-samples_cello}, we can find that our predicted heatmap are more synchronized with the motions of ground truth and our synthesized coarse video contains the basic texture and the poses. The final generated high-quality video has the comparable quality to the ground truth. 


\begin{table}[tb]
	\centering
	\caption{Quantitative evaluation of the proposed performance video generation with state-of-the-arts. }
	\label{tbl:video-comparsion}
	\begin{tabular}{lcccc} 
		\hline
		\multicolumn{1}{c}{\multirow{2}{*}{Methods}} & \multicolumn{2}{c}{Cello} & \multicolumn{2}{c}{Trombone} \\ 
		\cline{2-5}
		\multicolumn{1}{c}{} & \multicolumn{1}{l}{PSNR} & \multicolumn{1}{l}{SSIM} & \multicolumn{1}{l}{PSNR} & \multicolumn{1}{l}{SSIM} \\ 
		\hline
		EBDN \cite{chan2018everybody} & 13.553 & 0.246 & 12.358 & 0.225 \\ 
		vid2vid \cite{wang2018video} & 13.284 & 0.331 & 9.600 & 0.204 \\ \hline
		Ours & \textbf{17.073} & \textbf{0.563} & \textbf{15.910} & \textbf{0.397} \\ 
		\hline
	\end{tabular}
\end{table}

\subsection{User Study}	
\begin{figure}
	\centering
	\includegraphics[width=1\linewidth]{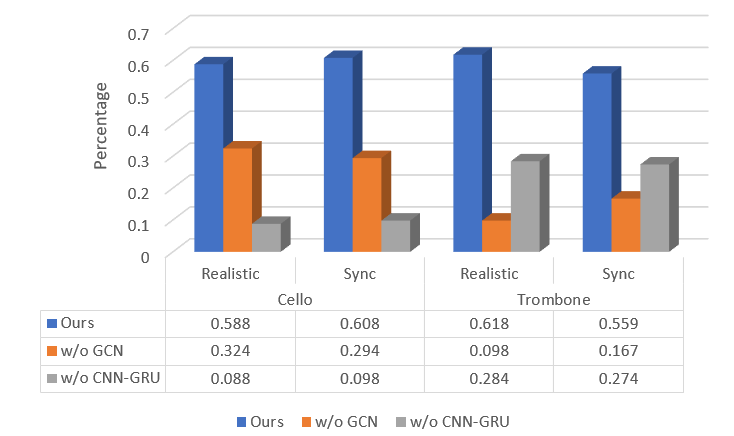}
	\caption{User study of our model and its variants on both realistic and synchronization (Sync).}
	\label{fig:userstudy}
\end{figure}

We further provide a user study together with two examples to demonstrate the effectiveness of our model in Fig. \ref{fig:supplymentarysamplecello} and Fig. \ref{fig:userstudy}. We first randomly select 12 video sets, each of which contains three videos generated by our method, our method without CNN-GRU and our method without GCN, then invite participants to vote on realistic and synchronism. Clearly, (1) our model achieves the highest rating than other variants in both realistic and synchronization. 
(2) Both the realistic and synchronization w/o CNN-GRU in Cello gain much lower rating than in Trombone. That means the CNN-GRU plays more important roles in Cello video generation by capturing the temporal consistency in slower motion videos (Cello). 
(3) GCN turns to play more important role in Trombone since the fast motion in videos (Trombone) affect less to keypoints prediction.

\section{Conclusion}
In this paper, we propose a novel multi-stage model for audio-driven performance video generation. To achieve this task, we first generate both global coarse video and local heatmap as middle information for final video generation. Then, we propose to transform keypoints to heatmap via a differentiable transforming function, since heatmap offers more spatial information while hard to generate from audio. Finally, a Structured Temporal UNet (STU) is designed to capture both intra-frame structured information via GCN module, and inter-frame temporal consistency via CNN-GRU based UNet module. Comprehensive experiments demonstrate the effectiveness of the proposed model.

\bibliographystyle{IEEEtran}
\bibliography{bare_conf}

%

\end{document}